\documentclass[10pt,twocolumn,letterpaper]{article}

\usepackage[pagenumbers]{cvpr} % To force page numbers, e.g. for an arXiv version

\definecolor{cvprblue}{rgb}{0.21,0.49,0.74}
\usepackage[pagebackref,breaklinks,colorlinks,allcolors=cvprblue]{hyperref}

\usepackage{url}                                 
\usepackage{graphicx}                                             
\newcommand{\bluetext}[1]{\textcolor{black}{#1}} 
\newcommand{\tibi}[1]{\textcolor{black}{#1}} 

\usepackage{booktabs} 
\usepackage{caption} % for captions
\usepackage{float} % for positioning  
\usepackage{tabularray}
\usepackage{multirow}
\usepackage{amsmath}
\usepackage{wrapfig}

\usepackage{amssymb}
\usepackage{fontawesome}
\usepackage[ruled,vlined]{algorithm2e} 
\usepackage{tikz}
\usetikzlibrary{calc}

\usepackage{pifont}
\usepackage{cuted}

\usepackage{enumitem} 
\usepackage{caption}
\usepackage{subcaption}
\usepackage[normalem]{ulem}
\usepackage{array}
\def\paperID{} % *** Enter the Paper ID here
\def\confName{CVPR}
\def\confYear{2026}

\title{FACT-GS: Frequency-Aligned Complexity-Aware Texture Reparameterization for 2D Gaussian Splatting}

\author{
Tianhao Xie\textsuperscript{1}\thanks{Equal contribution}\quad
Linlian Jiang\textsuperscript{1,2}\footnotemark[1]\quad
Xinxin Zuo\textsuperscript{1}\quad
Yang Wang\textsuperscript{1,2}\quad
Tiberiu Popa\textsuperscript{1} \\
\textsuperscript{1}Concordia University, Montréal, Canada \quad
\textsuperscript{2}Mila–Quebec AI Institute, Montréal, Canada \\
{\tt\small \{tianhao.xie,linlian.jiang\}@mail.concordia.ca} \\
{\tt\small \{xinxin.zuo,yang.wang,tiberiu.popa\}@concordia.ca}
}

\begin{document}

%\twocolumn[{%
%\renewcommand\twocolumn[1][]{#1}%
\maketitle
\begin{strip}
    \centering
    \vspace{-20mm}
    \includegraphics[width=\linewidth]{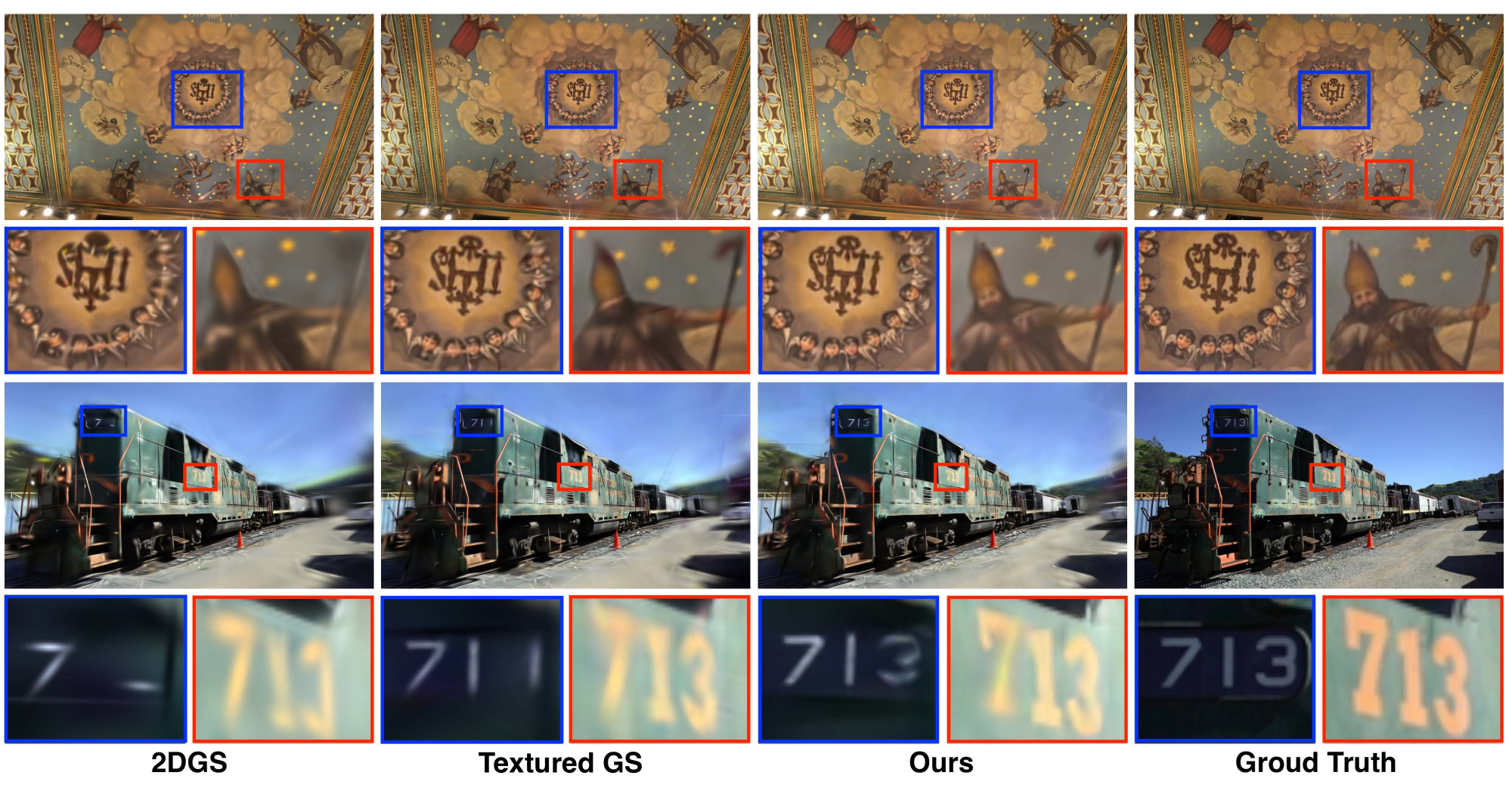}

    \captionof{figure}{
        Existing methods for novel view synthesis, such as 2DGS~\cite{huang20242d},
        use a spatially constant per-Gaussian appearance, while Textured GS~\cite{chao2025textured}
        adds per-Gaussian textures but still relies on a uniform sampling grid.
        This \textbf{\textit{uniform}} allocation ignores local signal complexity,
        causing high-frequency details to blur and wasting capacity in flat regions.
        In contrast, our \textbf{\textit{frequency-aligned}} texture reparameterization
        allocates capacity based on visual complexity, preserving sharp details 
        under the same primitive budget.
    }
    \label{fig:teaser}
    \vspace{2mm}
\end{strip}
\begin{abstract}
% \vspace{1em}
Realistic scene appearance modeling has advanced rapidly with Gaussian Splatting, which enables real-time, high-quality rendering. Recent advances introduced per-primitive textures that incorporate spatial color variations within each Gaussian, improving their expressiveness.
%using differentiable Gaussian primitives.
% However, per-Gaussian appearance in Gaussian-based representations %is parameterized using uniformly sampled textures,
% However, per-Gaussian appearance in Gaussian-based representations is parameterized uniformly in the space, allocating equal sampling density regardless of local visual complexity.
However, texture-based Gaussians parameterize appearance with a uniform per-Gaussian sampling grid, 
allocating equal sampling density regardless of local visual complexity, which leads to inefficient texture space utilization.
%This leads to inefficient texture space utilization, 
%where high-frequency regions are under-sampled and smooth regions waste capacity, causing blurred appearance and loss of fine structural detail.
% We introduce \textbf{FACT-GS}, a \textbf{\underline{F}}requency-\textbf{\underline{A}}ligned \textbf{\underline{C}}omplexity-Aware \textbf{\underline{T}}exture Gaussian Splatting framework that allocates sampling density according to local visual frequency.
% Grounded in adaptive sampling theory, FACT-GS reformulates texture parameterization as a differentiable sampling-density modulation problem, implemented via a learnable deformation field whose Jacobian determinant controls local sampling density.
We introduce \textbf{FACT-GS}, a \textbf{\underline{F}}requency-\textbf{\underline{A}}ligned \textbf{\underline{C}}omplexity-Aware \textbf{\underline{T}}exture Gaussian Splatting framework that allocates texture sampling density according to local visual frequency.
Grounded in adaptive sampling theory, FACT-GS reformulates texture parameterization as a differentiable sampling-density allocation problem, replacing the uniform textures with a learnable frequency-aware allocation strategy implemented via a
deformation field whose Jacobian modulates local sampling density.
Built on 2D Gaussian Splatting, FACT-GS performs non-uniform sampling on fixed-resolution texture grids,
preserving real-time performance while recovering sharper high-frequency details under the same parameter budget.
%To the best of our knowledge, this is the first frequency-adaptive appearance parameterization framework for Gaussian-based rendering.
%\vspace{-1em}
\end{abstract}

\vspace{-2.15em}
\section{Introduction}
\label{sec:intro}
Realistic 3D appearance modeling is essential for large-scale reconstruction, dynamic scene understanding, and photorealistic rendering~\citep{kulhanek2024wildgaussians,lin2024vastgaussian,fischer2024dynamic}, but achieving high fidelity under real-time constraints remains challenging.
Implicit volumetric fields such as NeRF~\citep{mildenhall2021nerf,barron2021mip,wu2024dynamic,zhao2023instant} provide high visual quality but require dense sampling and costly MLP inference, limiting scalability.
In contrast, 3D Gaussian Splatting (3DGS)~\citep{kerbl20233d,yu2024mip,feng2025flashgs} represents scenes using explicit differentiable Gaussian primitives, achieving real-time high-quality rendering.
Building on this, 2D Gaussian Splatting~\citep{huang20242d} uses planar primitives that simplify ray–primitive interaction and improve geometric stability.
% However, both 2D and 3D Gaussian splatting still rely on low-order spherical harmonics, which inherently encode only low-frequency appearance, thereby smoothing out fine textures, edges, and high-frequency reflectance.
%However, 2D and 3D Gaussian splatting represent appearance using low-order spherical harmonics~\citep{kerbl20233d}, which are inherently limited to low-frequency signals~\citep{ramamoorthi2001efficient,sloan2023precomputed}. This smooths out fine textures and high-frequency reflectance.
However, 2D and 3D Gaussian splatting represent appearance using view-dependent spherical harmonics~\citep{kerbl20233d,huang20242d}, without spatial color variation within each primitive, constraining its expressiveness.

To address this limitation, texture-based Gaussian extensions~\citep{chao2025textured,rong2025gstex,huang2024textured} attach per-Gaussian spatially varying textures. 
%providing a richer appearance representation and enabling detailed local variation for high-fidelity rendering. 
\tibi{However, the appearance parameterization of these methods is uniformly distributed across the texture space of each Gaussian, without adapting to local visual complexity~\citep{hachisuka2008multidimensional,sztrajman2021neural,wang2011efficiency}.
}
This uniform allocation leads to inefficient utilization of the texture space: high-frequency regions (e.g., sharp edges, fine patterns, numerical markings) are allocated insufficient texture space, resulting in detail loss, while large smooth regions inefficiently consume capacity to represent nearly uniform color. In practice, each Gaussian can only store a low-resolution texture patch (e.g., $4\times4\times4$), exacerbating the expressiveness constraints imposed by uniform sampling.
As shown in Fig.~\ref{fig:teaser}, this results in blurred appearances in visually complex regions.
%and unnecessary redundancy in smooth regions.
Simply increasing texture resolution yields only marginal improvements at quadratically increased memory and bandwidth cost, while neural texture fields~\citep{xu2024texture,zhang2025neural}
%, although more expressive, 
break the real-time rendering performance \tibi{of the} Gaussian-based representations.

%Despite these advances, the appearance parameterization in Gaussian-based representations~\citep{chao2025textured,rong2025gstex,huang2024textured} still suffers from a \textbf{structural inefficiency}: 
%sampling capacity is allocated uniformly across texture space, without adapting to local visual complexity~\citep{hachisuka2008multidimensional,sztrajman2021neural,wang2011efficiency}.
%This uniform parameterization treats all texels equally, regardless of their visual importance, leading to a \textbf{sampling–complexity mismatch}
%—a structural issue in texture mapping and adaptive sampling where a spatially uniform sampling grid is decoupled from the underlying signal frequency~\citep{heckbert1989fundamentals,rousselle2011adaptive,zwicker2015recent}.

%In practice, each Gaussian stores a fixed-resolution texture patch, enforcing this uniform sampling and causing high-frequency regions (e.g., sharp edges, fine patterns, numerical markings) to receive too few samples (detail loss), while large smooth regions waste capacity encoding nearly constant color.

% OLD
%Motivated by this structural inefficiency, we reformulate the per-Gaussian texture appearance as an adaptive sampling problem: sampling should follow the local signal frequency rather than remain spatially uniform. 
\tibi{
Motivated by this structural inefficiency, we propose an adaptive sampling scheme of the per-Gaussian texture where sampling should follow the local signal frequency rather than remain spatially uniform. 
}
To achieve this, we introduce a frequency-aligned differentiable texture reparameterization that continuously modulates
sampling density across the texture domain. 
Grounded in adaptive sampling theory~\citep{heckbert1989fundamentals,rousselle2011adaptive,zwicker2015recent},
this allocates more sampling capacity to high-frequency regions and less to smooth areas. 
Building on this formulation, we present \textbf{FACT-GS}, a Frequency-Aligned Complexity-Aware Texture Gaussian Splatting framework that integrates this frequency-adaptive reparameterization. %In essence, it replaces uniform texture layouts with a learnable, frequency-aware allocation rule, improving the efficiency of texture space utilization by information-driven sampling.
%shifting appearance modeling from geometry-driven to information-driven sampling.

% our method replaces uniform texture parameterization with a dynamic, frequency-aware allocation rule, shifting appearance modeling from geometry-driven layouts to information-driven sampling.

% Grounded in classical adaptive sampling theory~\citep{heckbert1989fundamentals,rousselle2011adaptive,zwicker2015recent}, our formulation establishes a direct correspondence between local signal frequency and optimal sampling density, enabling textures to self-adjust their effective resolution where detail is needed.

Concretely, FACT-GS is built upon 2D Gaussian Splatting~\citep{huang20242d} and introduces a learnable deformation field defined in the texture domain. 
This field predicts continuous per-texel displacements, inducing a smooth warping whose Jacobian determinant~\citep{sander2001texture} directly modulates local sampling density. 
In contrast to previous uniform texture parameterizations~\citep{chao2025textured,rong2025gstex,huang2024textured},
which allocate equal resolution regardless of signal complexity, our warping adaptively concentrates samples in high-frequency regions while compressing low-frequency areas. 
This realizes adaptive sampling as a differentiable, per-primitive reparameterization, allowing texture capacity to naturally focus where visual detail is needed. 
The entire framework is trained end-to-end under standard photometric supervision, 
as the deformation field is designed to integrate seamlessly into the differentiable splatting pipeline.
As a result, FACT-GS recovers sharper high-frequency
appearance and uses parameters more efficiently, all while maintaining real-time rendering performance under the same primitive budget.

Our contributions are summarized as follows:

\begin{itemize}
\item We propose FACT-GS, a frequency-aligned texture reparameterization that reallocates sampling density based on local signal frequency under a fixed texture budget.

\item We reformulate the texturing problem through the lens of adaptive sampling theory as a sampling-density allocation problem, achieved via a learnable deformation field whose Jacobian controls local texture space distortion.

% Our method improves the texture parameterization by employing an adaptive sampling scheme that optimizes the sampling density of the texture via a learnable deformation field whose Jacobian controls local area distortion

%\item We cast texture parameterization as a differentiable sampling-density allocation via a learnable deformation field whose Jacobian controls local area distortion.

%We provide an analysis underpinning our method, grounded in the principles of adaptive sampling.

\item We demonstrate through extensive qualitative, quantitative experiments, and ablation studies that our method provides consistent, detail-preserving improvements over state-of-the-art textured Gaussian methods across benchmarks without sacrificing real-time performance.
\end{itemize}

\section{Related Work}
\label{sec:relatedwork}
\noindent\textbf{Novel View Synthesis.} 
Novel view synthesis (NVS) aims to generate photorealistic renderings of 3D scenes from unseen viewpoints, typically given multi-view images with known camera poses. 
Its development originates from classical SfM and MVS pipelines~\citep{snavely2006photo,snavely2008modeling,agarwal2011building,schonberger2016structure,seitz2006comparison,furukawa2009accurate,furukawa2010towards}, 
which reconstruct geometry through multi-view correspondences but suffer from occlusion sensitivity and limited viewpoint extrapolation.  
Neural radiance fields (NeRF)~\citep{mildenhall2021nerf} later revolutionized NVS by representing scenes as implicit neural functions optimized via photometric supervision. 
Subsequent extensions improve anti-aliasing~\citep{barron2021mip}, large-scale reconstruction~\citep{tancik2022block,barron2022mip}, reflection modeling~\citep{verbin2022ref}, and acceleration via grid-based encodings~\citep{zhao2023instant}, 
yet they remain constrained by dense ray sampling and slow MLP inference.

This motivates explicit representations such as 3D Gaussian Splatting~\citep{kerbl20233d}, which replace neural volumes with differentiable Gaussian primitives for real-time rendering. 
Building upon this foundation, follow-up works improve geometric accuracy and rendering stability through enhanced kernel formulations and densification strategies~\citep{fu2024colmap,lei2025mosca,li20253d}. 
To further refine geometry, 2D Gaussian Splatting~\citep{huang20242d} introduces planar Gaussian primitives for more accurate and efficient radiance field representation, enabling fine-grained geometric reconstruction while maintaining high appearance quality. However, both 3D and 2D Gaussian Splatting modeled appearance via view-dependent spherical harmonic coefficients, but lack spatial color variation within each primitive, limiting their expressiveness.
%the low-order spherical-harmonic color encoding in 2D or 3D Gaussians limits its ability to capture high-frequency appearance details. 

\noindent\textbf{Spatially Varying Gaussians.}
Recent works introduce spatially varying Gaussians to overcome this constraint. 
One line of research augments each primitive with RGB or RGBA textures to model spatially varying colors~\citep{chao2025textured,xu2024supergaussians,rong2025gstex,huang2024textured,song2024hdgs} 
or surface normals for relighting~\citep{sun2025svg}. 
Another employs neural texture fields to decouple geometry and appearance~\citep{xu2024texture,zhang2025neural}, achieving high visual fidelity at the cost of slower training and rendering due to MLP inference.

Texture-based Gaussians strike a balance between quality and efficiency, 
providing stronger appearance modeling without sacrificing real-time performance. 
However, their uniformly sampled texture parameterization remains inefficient, 
failing to allocate texture capacity according to signal complexity.
% We address this limitation through a frequency-aligned reparameterization 
% that adaptively reallocates sampling density according to local signal complexity, concentrating texture capacity where visual detail is needed.
This uniform sampling strategy leads to a structural sampling–complexity mismatch,
where high-frequency regions receive insufficient capacity while smooth regions waste parameters.

%%% Note: cite not enough
\noindent\textbf{Adaptive Sampling and Frequency-Aware Parameterization.}
Classical rendering and Monte Carlo integration extensively explored adaptive sampling~\citep{hachisuka2008multidimensional,zwicker2015recent,rousselle2011adaptive,sztrajman2021neural,wang2011efficiency}, 
where sample density is modulated by local signal frequency or reconstruction error to achieve efficient high-frequency reconstruction.
At the representation level, multi-resolution texture filtering such as mipmapping and anisotropic filtering~\citep{heckbert2007survey,williams1983pyramidal,pharr2024filtering}
pre-filters textures across scales to suppress aliasing and maintain visual consistency under varying viewing distances.
Recent neural and Gaussian-based formulations~\citep{li2024dngaussian,zhu2024fsgs,liu20243dgs}
extend this principle to learned 3D representations, enforcing frequency- or scale-consistent parameterization in both geometry and appearance reconstruction.
However, these methods operate at the global or feature-map level and still assume uniform sampling within each primitive,
limiting their ability to adapt texture-space resolution to local signal variation.

To address this, we introduce a differentiable frequency-aligned reparameterization
that operates at the per-primitive coordinate level,
bridging adaptive sampling theory with texture parameterization in Gaussian representations.

\section{Preliminaries}
\label{sec:preliminaries}
\begin{figure*}
    \centering
    \includegraphics[width=\linewidth]{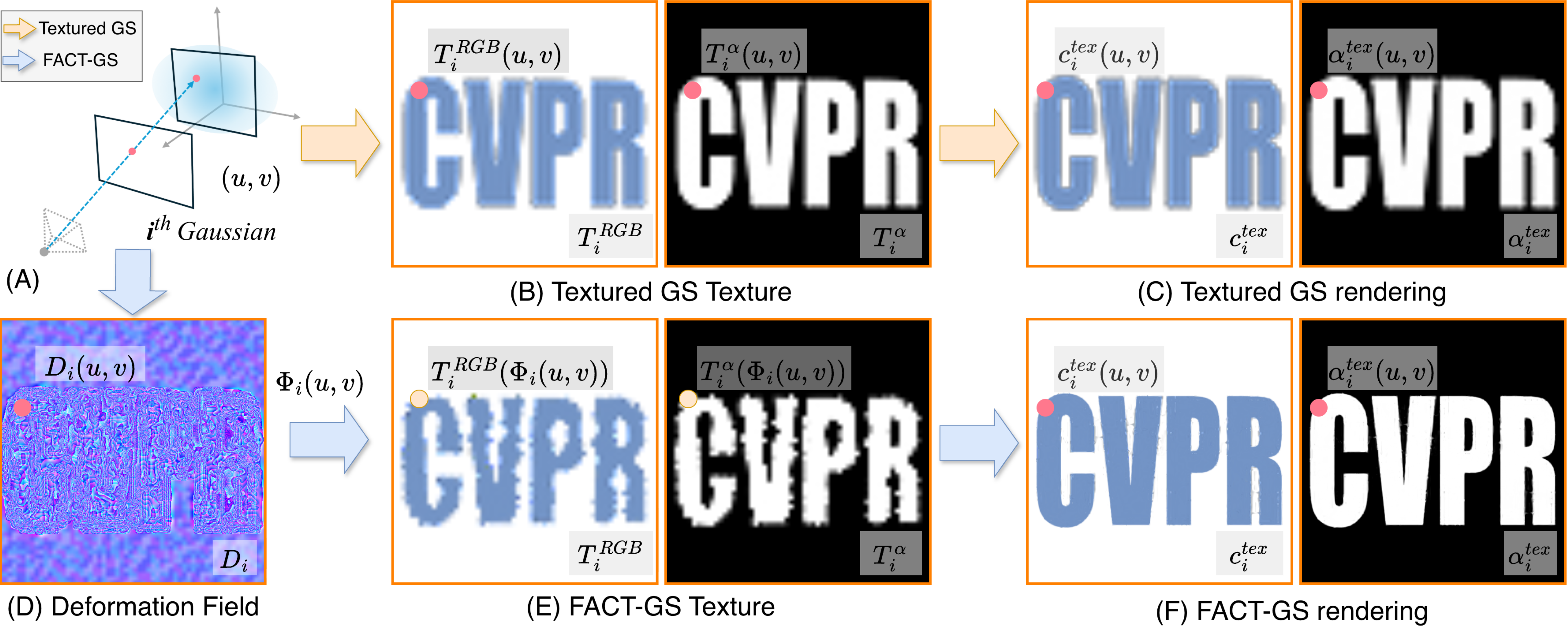}
    \caption{Method comparison of Textured GS and FACT-GS. (A) For the $i$-th Gaussian, the ray–Gaussian intersection was firstly computed to obtain texture coordinates $(u,v)$. Textured GS bilinearly sampled from RGB and opacity textures $\mathbf{T}_i^{\text{RGB}}$ and $\mathbf{T}_i^{\alpha}$ (B) using $(u,v)$ to get the texture color $c_i^{\text{tex}}$ and opacity $\alpha_i^{\text{tex}}$ (C), which lead to blurred high-frequency details (edges). Instead, FACT-GS uses a learnable deformation field $\mathbf{D}_i$ to predict a continuous warp $\Phi_i(u,v)$ that allocate sampling density guided by local frequency (D), after which the warped coordinates $\Phi_i(u,v)$ are used to sample from RGB and opacity textures (E), producing the FACT texture color and opacity (F), which has more clear high-frequency details.}
    \label{fig:method_overview}
\end{figure*}
\subsection{2D Gaussian Splatting Representation}
\label{sec:2drep}

We adopt the 2D Gaussian Splatting representation~\cite{huang20242d}. 
Each primitive lies on a local tangent plane with center $p_k$, orthonormal basis $(t_\beta,t_\gamma,n_k)$, 
and scales $(s_\beta,s_\gamma)$. Local coordinates on the plane are $(\beta,\gamma)$:
\begin{equation}
P(\beta,\gamma)=p_k+s_\beta t_\beta\beta+s_\gamma t_\gamma\gamma.
\label{eq:localparam}
\end{equation}
The spatial weight is $\mathcal{G}(\beta,\gamma)=\exp[-(\beta^2+\gamma^2)/2]$, and splats are projected and
composited in image space. 
Appearance is modeled using spherical harmonics, producing one view-dependent color per primitive,
without spatial variation across the tangent plane.

\subsection{Textured Gaussians}
\label{sec:texgs}

Standard 2D Gaussians provide only a single view-dependent color for each primitive. 
Textured Gaussians~\cite{chao2025textured} instead attach a learnable RGBA texture map 
$\mathbf{T}_i \in \mathbb{R}^{\tau \times \tau \times 4}$ to each primitive, enabling spatially varying 
appearance within tangent plane. 
Although originally formulated in 3D, the official implementation adopts a 2D variant for more accurate 
and efficient ray–Gaussian intersection, which we also follow.

Given the local intersection coordinates $\mathbf{s}(\mathbf{x})=(\beta,\gamma)$ on the tangent plane, the corresponding texture coordinates $(u,v)$ are computed as
% \vspace{-2pt}
\begin{equation}
u=\frac{\beta+\xi}{2\xi}\tau,\quad
v=\frac{\gamma+\xi}{2\xi}\tau,
\label{eq:uv_map}
\end{equation}
where $\xi$ controls the effective texture support.
The texture color $c_i^{\mathrm{tex}}(\mathbf{p})$ and opacity $\alpha_i^{\mathrm{tex}}(\mathbf{p})$ 
are bilinearly sampled from the texture map $\mathbf{T}_i(u,v)$.
We combine these with the SH-encoded base color $c_i^{\mathrm{base}}(\mathbf{p})$ and base opacity $o_i$, 
and composite the final color using volumetric alpha blending weighted by the Gaussian footprint 
$\mathcal{G}_i(\mathbf{s}(\mathbf{x}))$:
\begin{equation}
\label{eq:tex3}
\begin{aligned}
c(\mathbf{p}) &=
\sum_{i=1}^{N}
\bigl(c_i^{\mathrm{tex}}(\mathbf{p}) + c_i^{\mathrm{base}}(\mathbf{p})\bigr)\,
\bigl(\alpha_i^{\mathrm{tex}}(\mathbf{p})\,\mathcal{G}_i(\mathbf{s}(\mathbf{x}))\,o_i \bigr) \\
&\quad\times
\prod_{j<i}
\Bigl[1 - \alpha_j^{\mathrm{tex}}(\mathbf{p})\,\mathcal{G}_j(\mathbf{s}(\mathbf{x}))\,o_j \Bigr].
\end{aligned}
\end{equation}

\section{Method}
\label{sec:method}
%Our method adapts texture sampling density to local appearance complexity. 
%We first identify that uniform texture parameterization leads to an inefficient texture space utilization (Sec.~\ref{sec:mismatch}),
%\textit{sampling–complexity mismatch} (Sec.~\ref{sec:mismatch}), 
%then reformulate texture mapping as an \textit{adaptive sampling} problem (Sec.~\ref{sec:allocate}),
%\textit{sampling-density allocation problem} (Sec.~\ref{sec:allocate}),
%and finally realize this principle via a \textit{frequency-aligned reparameterization} implemented as a learnable deformation field $\Phi$ (Sec.~\ref{sec:ourmethod}).

\subsection{Motivation}
 %Sampling-Complexity Mismatch
\label{sec:mismatch}

While Textured Gaussians introduce spatial appearance variation through learnable texture maps $\mathbf{T}_i$,
these textures are parameterized on a uniform grid that allocates equal representational capacity within each splat regardless of local signal frequency.
As a result, high-frequency regions that occupy small spatial areas are under-represented, whereas smooth regions that span larger areas overconsume parameters to represent redundant information, leading to a persistent sampling–complexity mismatch in the texture domain and imbalanced texture capacity allocation.

This mismatch manifests in two characteristic failure modes:
\textbf{(1) Detail loss.} High-frequency regions are assigned too few texels, resulting in insufficient representational capacity to capture fine details (see Fig.~\ref{fig:teaser}; quantified in Fig.~\ref{fig:tex_compare}).
\textbf{(2) Inefficient resolution scaling.} Increasing texture resolution raises memory and bandwidth quadratically, yet provides only marginal visual improvement since the sampling pattern remains uniform.

Fundamentally, this imbalance reflects a disconnect between texture capacity 
and the spatial frequency of the underlying appearance, motivating the need 
for frequency-aware allocation.
%\subsection{Texture as a Deformable Sampling Space}
%\label{sec:allocate}
To address the sampling--complexity mismatch in texture space, we reinterpret texture parameterization as a sampling-density allocation problem through the lens of adaptive sampling theory.

Let $C(u,v)$ denote the ground-truth color function, and let the texture $\mathbf{T}$ be its discrete representation, where texels sample $C(u,v)$ in texture space. The sampling density $\rho(u,v)$ represents the number of texels 
per unit area. Uniform parameterization is structurally inefficient because it 
assigns the same density everywhere despite large spatial variation in the 
underlying appearance signal~\citep{sander2001texture,zwicker2015recent}.

Classical reconstruction theory suggests that sampling density should increase in 
regions where the signal varies more rapidly. Since a texture map is a discrete 
sampling of the continuous appearance function $C(u,v)$, its local variation 
behaves analogously to 2D image signals in reconstruction theory: higher spatial 
variation corresponds to higher local frequency and therefore requires denser 
sampling.

Following prior work that uses local variation as a proxy for frequency
content~\citep{eskicioglu2002image,rousselle2011adaptive}, we adopt the
gradient-based target density
\begin{equation}
\rho^\star(u,v) \propto (\|\nabla C(u,v)\| + \epsilon)^{\alpha},
\label{eq:optimal_density}
\end{equation}
where $\|\nabla C\|$ estimates local spatial frequency, $\alpha$ controls
adaptation strength, and $\epsilon$ ensures stability.

Under a continuous reparameterization $\Phi$, a region around $(u,v)$
\tibi{in the parameter space} with area $dA$ is mapped to \tibi{a region with area} $|\det J_{\Phi}(u,v)|\, dA$ in warped texture space, \tibi{therefore}
%Since texels are uniformly distributed in the warped domain, 
the induced
sampling density is modulated by $|\det J_{\Phi}(u,v)|$.
%\begin{equation}
%\rho_{\Phi}(u,v) \propto |\det J_{\Phi}(u,v)|.
%\end{equation}
Thus, a larger $|\det J_{\Phi}|$ indicates higher local texture capacity.
Matching $\rho^\star(u,v)$ therefore amounts to shaping the Jacobian
determinant of $\Phi$. Because photometric loss concentrates gradients in
high-frequency regions, optimization naturally expands $\Phi$ in these areas,
increasing $|\det J_{\Phi}|$ until the induced density aligns with $\rho^\star$.

\bluetext{These observations motivate treating texture space as a deformable sampling domain whose density adapts to local visual complexity. We next introduce a frequency-aligned reparameterization that realizes this idea.}

\subsection{Frequency-Aligned Complexity-Aware Texture Reparameterization}
\label{sec:ourmethod}

We realize the target sampling density $\rho^\star(u,v)$ by introducing
a learnable warp $\Phi:(u,v)\mapsto(u',v')$ over texture space.
By reshaping the domain, the warp redistributes texels according to
signal complexity: high-frequency regions are expanded to receive more
texels, while smooth regions are compressed to avoid redundant sampling.
This redistribution is realized through the local area change of the warp,
captured by the Jacobian determinant $|\det J_{\Phi}(u,v)|$.

In contrast to texture-based Gaussians~\citep{chao2025textured,rong2025gstex},
which operate on fixed uniform grids and primarily improve detail by
increasing resolution, our formulation adjusts sampling density
geometrically through $\Phi$ while keeping the grid resolution fixed.
This enables non-uniform, complexity-aware texture parameterization
that remains fully differentiable and end-to-end trainable.

\textbf{(i) Differentiable Warping Field.}
For each Gaussian, we instantiate $\Phi$ using a deformation field
$\mathbf{D}_i \in \mathbb{R}^{\tau\times\tau\times2}$ that predicts
per-texel displacements,
$\mathbf{D}_i(u,v) = (D_i^u(u,v), D_i^v(u,v))$.
Warped coordinates are obtained as
\begin{equation}
(u',v') = (u,v) + \lambda\,\mathbf{D}_i(u,v),
\label{eq:warp}
\end{equation}
defining the continuous mapping $\Phi_i$.
The Jacobian $J_{\Phi_i}$ captures the resulting local area distortion
and therefore the effective sampling density at each texel.

\textbf{(ii) Texture Sampling and Gradient Modulation.}
The warped coordinates $\Phi_i(u,v)$ are used for differentiable texture lookup:
\begin{equation}
c_i^{\mathrm{tex}}(\mathbf{p})=\mathbf{T}_i^{\mathrm{RGB}}\!\big(\Phi_i(u,v)\big),\quad
\alpha_i^{\mathrm{tex}}(\mathbf{p})=\mathbf{T}_i^{\alpha}\!\big(\Phi_i(u,v)\big).
\label{eq:pixel_color}
\end{equation}

Differentiating Eq.~(\ref{eq:pixel_color}) with respect to $(u,v)$ and
applying the chain rule yields:
\begin{equation}
\nabla c_i^{\mathrm{tex}}(u,v)
=
J_{\Phi_i}(u,v)^{\!\top}\,
\nabla \mathbf{T}_i^{\mathrm{RGB}}\!\big(\Phi_i(u,v)\big).
\end{equation}
Here, $\nabla \mathbf{T}_i^{\mathrm{RGB}}$ denotes the spatial gradient
of the uniform-grid texture. This expression shows that the warp
reshapes the texture’s local gradient field through $J_{\Phi_i}^{\top}$,
effectively modulating local frequency content.

%The Jacobian of the warp in Eq.~(\ref{eq:warp}) is
%\begin{equation}
%J_{\Phi_i}(u,v)=
%\begin{bmatrix}
%1+\lambda\,\partial_u D_i^u & \lambda\,\partial_v D_i^u\\
%\lambda\,\partial_u D_i^v & 1+\lambda\,\partial_v D_i^v
%\end{bmatrix},
%\label{eq:jacobian}
%\end{equation}
%which characterizes the local area change and thus the induced
%sampling-density modulation.

The local signal frequency can be approximated by the gradient magnitude
$\|\nabla c_i^{\mathrm{tex}}(u,v)\|$~\cite{eskicioglu2002image}. 
Under the warp, the uniform-grid texture gradient 
$\nabla \mathbf{T}_i^{\mathrm{RGB}}$ is transformed by the singular values 
of $J_{\Phi_i}$, which respectively amplify or attenuate its magnitude. 
These effects manifest as regions receiving more texels allocated
or fewer texels received, enabling the FACT texture 
$c_i^{\mathrm{tex}}(u,v)$ to respond faithfully to higher-frequency
variation in the ground-truth color function $C(u,v)$.

This establishes the connection between geometric reparameterization and
frequency-aligned density allocation. The deformation field
$\mathbf{D}_i$ is trained end-to-end via the photometric loss, which
implicitly drives $|\det J_{\Phi_i}(u,v)|$ toward the target density
$\rho^\star(u,v)$.

\textbf{(iii) Final Rendering.}
After warping, the per-Gaussian colors and opacities are evaluated at
$\Phi_i(u,v)$ and composited using the standard differentiable
splatting procedure in Eq.~(\ref{eq:tex3}).
The rendering pipeline itself remains unchanged; only the texture
parameterization is modified, ensuring that real-time Gaussian
splatting is fully preserved.

\bluetext{The complete FACT-GS pipeline is illustrated in
Fig.~\ref{fig:method_overview}, and Alg.~\ref{alg:FACTr} summarizes
the training process.}

\begin{algorithm}[t]
\LinesNumbered
\caption{Training of FACT-GS}
%\caption{Frequency-Aligned Complexity-Aware Texture Reparameterization}
\label{alg:FACTr}
\KwIn{Pretrained Gaussian parameters $\{p_i, s_\beta, s_\gamma, t_\beta, t_\gamma, o_i, \mathrm{SH}_i\}$, textures $\{\mathbf{T}_i\}$ and deformation field $\{\mathbf{D}_i\}$}
\KwOut{Rendered image $\hat{I}$ with aligned details}

\BlankLine
%\textbf{Training:}\\
\For{each Gaussian $i=1,\dots,N$}{
    Compute $(u,v)$ via tangent-plane mapping;\\
    Predict displacement $\mathbf{D}_i(u,v)$;\\
    Warp coordinates $(u',v')=(u,v)+\mathbf{D}_i(u,v)$;\\
    Sample $c_i^{\mathrm{tex}}, \alpha_i^{\mathrm{tex}}$ from $\mathbf{T}_i(u',v')$;\\
}

Blend all Gaussians via Eq.~(\ref{eq:tex3}) to get $\hat{I}$;\\
Backpropagate loss to update all parameters.

\BlankLine
%\textbf{Inference:}\\
%Render with learned deformation $\Phi_i$ in a single forward pass.
\end{algorithm}

\begin{figure*}[t]
    \centering
    \includegraphics[width=\linewidth]{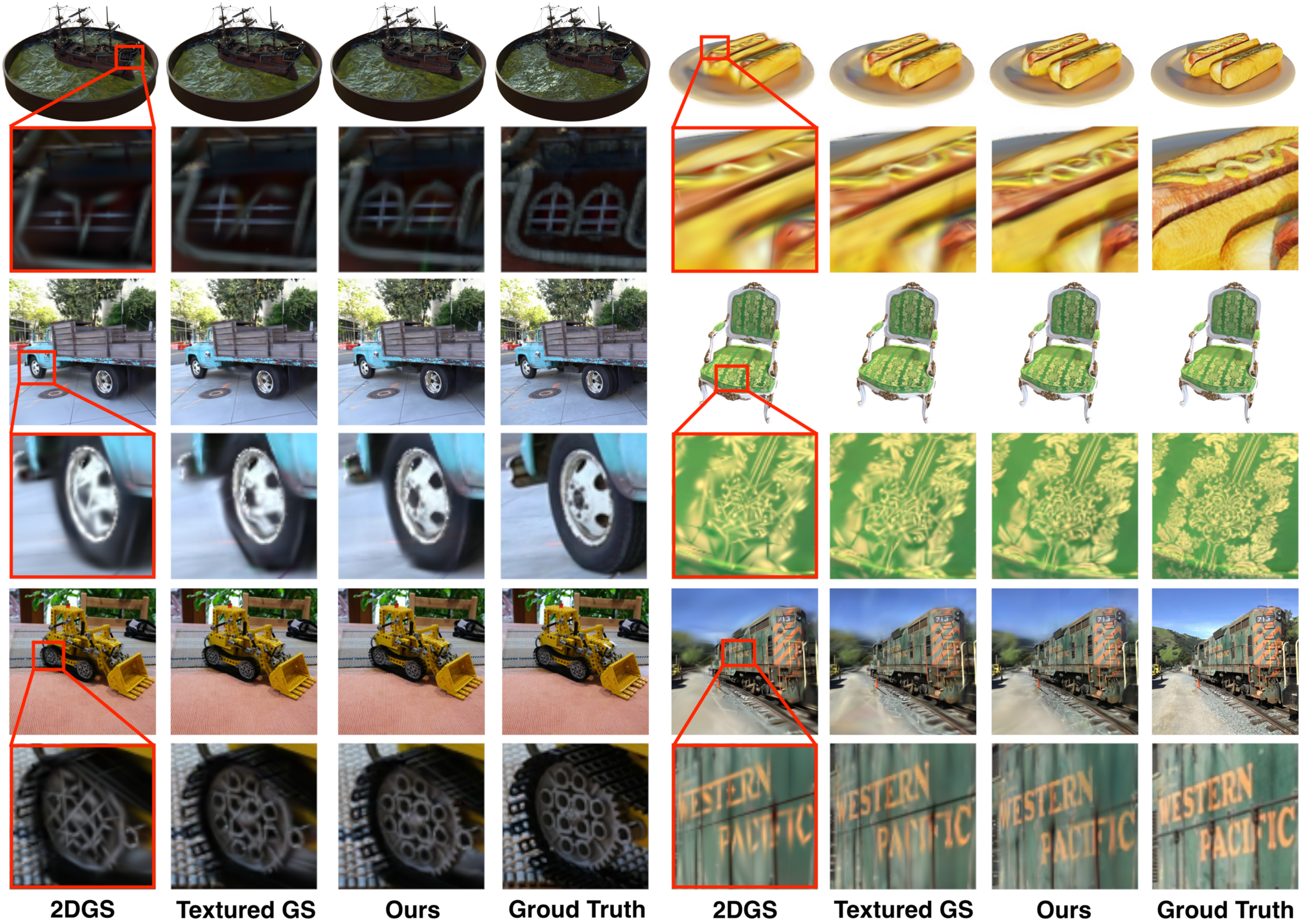}
\caption{
Novel view synthesis comparison under reduced primitive budgets (10\% / 1\% of default 2DGS, exact budgets are shown in Supp.~\ref{sec:budgets}).
2DGS~\citep{huang20242d} loses high-frequency appearance, and Textured GS~\citep{chao2025textured} blurs fine detail due to uniform texture sampling.
Our method preserves sharp edges, textures, and material details (zoom-ins),
achieving significantly higher perceptual fidelity without increasing parameters.
}
    \label{fig:results_few}
\end{figure*}

\noindent\textbf{Discussion.}
Our formulation reframes texture parameterization as an adaptive sampling problem: the objective is no longer to uniformly cover the texture domain, but to allocate representational capacity in proportion to local signal frequency.
%under a fixed parameter budget
The continuous mapping $\Phi$ provides a learnable mechanism that moves the effective sampling density toward the frequency-aligned target, shifting texture layouts from geometry-driven to information-driven organization.

Although instantiated on Gaussian textures, the principle is general:
any spatially parameterized appearance representation (e.g., feature planes, neural textures, volumetric grids) can adopt frequency-aligned sampling to improve the distribution of representational capacity.

In essence, FACT-GS replaces uniform parameterization with a dynamic,
frequency-aware allocation rule, enabling higher visual fidelity under the same parameter budget.

\subsection{Optimization}

Following Textured Gaussians~\citep{chao2025textured}, 
we adopt a two-stage optimization for stability.  
In the first stage, the 2D Gaussian primitives are trained using the standard losses from 3D Gaussian Splatting~\citep{kerbl20233d}, 
combining $\mathcal{L}_1$ photometric and structural similarity ($\mathcal{L}_{\text{SSIM}}$) terms, 
augmented with a mask loss $\mathcal{L}_{\alpha}$ when ground-truth masks are available: 
\begin{equation}
\mathcal{L} = \eta \mathcal{L}_1 + (1-\eta)\mathcal{L}_{\text{SSIM}} + \mathcal{L}_{\alpha},\quad \eta = 0.2.
\end{equation}
This stage runs for $30{,}000$ iterations.  
In the second stage, we optimize the per-Gaussian texture $\mathbf{T}$ 
and deformation field $\mathbf{D}$ 
while fine-tuning the Gaussian parameters 
$\{p_k, s_\beta, s_\gamma, t_\beta, t_\gamma, o, \text{SH}\}$ 
without pruning or densification, 
for another $30{,}000$ iterations.  
Learning rates for $\mathbf{T}$ and $\mathbf{D}$ 
are set to $2.5\times10^{-3}$ and $1\times10^{-3}$, respectively.

\section{Experiments}
\subsection{Implementation Details}
We build on the gsplat~\citep{ye2025gsplat} framework and extend it with a
lightweight CUDA kernel for frequency-aligned texture warping. The per-Gaussian texture space sampling-density modulation described in section~\ref{sec:ourmethod} introduces negligible overhead during inference. All experiments are run on
a single NVIDIA RTX A6000 GPU.

% As shown in Table~\ref{tab:inference_speed}, our renderer runs within $1\text{–}2$ FPS of Textured GS~\citep{chao2025textured} on NeRF Synthetic, while preserving real-time performance on a single NVIDIA RTX A6000 GPU.

\subsection{Datasets and Evaluation Metrics}
\noindent\textbf{Dataset. }We evaluate our method on five standard benchmarks for novel view synthesis: 
NeRF Synthetic~\citep{mildenhall2021nerf} (8 scenes), 
MipNeRF 360 v2~\citep{barron2022mip} (7 scenes), 
DTU~\citep{jensen2014large} (4 scenes), 
Tanks \& Temples~\citep{knapitsch2017tanks} (3 scenes), 
and LLFF~\citep{mildenhall2019llff} (2 scenes).

\noindent\textbf{Metrics.} Following Textured GS~\cite{chao2025textured}, we report Peak Signal-to-Noise Ratio (PSNR), Structural Similarity Index Measure (SSIM), and Learned Perceptual Image Patch Similarity (LPIPS)~\cite{zhang2018unreasonable} for quantitative evaluation.

\subsection{Experimental Results}

\setlength{\tabcolsep}{0.765mm} 
\begin{table*}[t]
\centering
\scriptsize
\setlength{\arrayrulewidth}{0.6pt}   % thicker rules
\renewcommand{\arraystretch}{1.22} % increase row spacing
% \resizebox{\linewidth}{!}{
\begin{tabular}{l|lcccl|lcccl|lcccl|lcccl|lcccl} 
\toprule[0.8pt]    % make top rule thicker
\multirow{2}{*}{\textbf{Methods}}   & \multicolumn{5}{c|}{\textbf{NeRF Synthetic}~\cite{mildenhall2021nerf}}                                      &  & \multicolumn{3}{c}{\textbf{MipNeRF 360v2}~\cite{barron2022mip}}                                                                              &  &  & \multicolumn{3}{c}{\textbf{DTU}~\cite{jensen2014large}}                                                                                        &  &  & \multicolumn{3}{c}{\textbf{Tanks \& Temples}~\cite{knapitsch2017tanks}}                                                                              &  &  & \multicolumn{3}{c}{\textbf{LLFF}~\cite{mildenhall2019llff}}                                                                                       &   \\ 
\cline{3-5}\cline{8-10}\cline{13-15}\cline{18-20}\cline{23-25}
                    &  & PSNR$\uparrow$ & SSIM$\uparrow$          & LPIPS$\downarrow$        &  &  & \multicolumn{1}{l}{PSNR$\uparrow$} & \multicolumn{1}{l}{SSIM$\uparrow$} & \multicolumn{1}{l}{LPIPS$\downarrow$} &  &  & \multicolumn{1}{l}{PSNR$\uparrow$} & \multicolumn{1}{l}{SSIM$\uparrow$} & \multicolumn{1}{l}{LPIPS$\downarrow$} &  &  & \multicolumn{1}{l}{PSNR$\uparrow$} & \multicolumn{1}{l}{SSIM$\uparrow$} & \multicolumn{1}{l}{LPIPS$\downarrow$} &  &  & \multicolumn{1}{l}{PSNR$\uparrow$} & \multicolumn{1}{l}{SSIM$\uparrow$} & \multicolumn{1}{l}{LPIPS$\downarrow$} &   \\ 
\hline
2D GS (100\%)       &  & 33.38          & 0.966                   & 0.0275                   &  &  & 28.96                              & 0.870                              & 0.0997                                &  &  & 27.85                              & \uline{0.905}                      & 0.1328                                &  &  & 22.79                              & 0.823                              & 0.1572                                &  &  & 27.32                              & 0.883                              & 0.1092                                &   \\
Textured GS (100\%) &  & \uline{33.91}  & \uline{0.968}           & \uline{0.0235}           &  &  & \uline{29.30}                      & \uline{0.873}                      & \uline{0.0915}                        &  &  & \textbf{28.76}                     & \uline{0.905}                      & \uline{0.1241}                        &  &  & \textbf{23.56}                     & \uline{0.831}                      & \uline{0.1428}                        &  &  & \textbf{29.04}                     & \uline{0.899}                      & \uline{0.0852}                        &   \\
Ours (100\%)        &  & \textbf{34.02} & \textbf{\textbf{0.969}} & \textbf{\textbf{0.0220}} &  &  & \textbf{29.34}                     & \textbf{\textbf{0.874}}            & \textbf{\textbf{0.0889}}              &  &  & \textbf{28.76}                     & \textbf{\textbf{0.906}}            & \textbf{\textbf{0.1204}}              &  &  & \uline{23.53}                      & \textbf{\textbf{0.834}}            & \textbf{\textbf{0.1377}}              &  &  & 28.99                              & \textbf{\textbf{0.900}}            & \textbf{\textbf{0.0827}}              &   \\ 
\hline
2D GS (10\%)        &  & 30.35          & 0.945                   & 0.0594                   &  &  & 27.22                              & 0.805                              & 0.1937                                &  &  & \textbf{29.04}                     & \uline{0.899}                      & 0.1791                                &  &  & 22.18                              & 0.767                              & 0.2577                                &  &  & 27.37                              & 0.863                              & 0.1491                                &   \\
Textured GS (10\%)  &  & \uline{30.88}  & \uline{0.949}           & \uline{0.0542}           &  &  & \uline{27.80}                      & \uline{0.825}                      & \uline{0.1646}                        &  &  & 28.72                              & 0.894                              & \uline{0.1640}                        &  &  & \uline{22.65}                      & \uline{0.792}                      & \uline{0.2206}                        &  &  & \uline{28.29}                      & \uline{0.884}                      & \uline{0.1206}                        &   \\
Ours (10\%)         &  & \textbf{31.51} & \textbf{\textbf{0.954}} & \textbf{\textbf{0.0439}} &  &  & \textbf{28.06}                     & \textbf{\textbf{0.837}}            & \textbf{\textbf{0.1447}}              &  &  & \uline{28.79}                      & \textbf{\textbf{0.904}}            & \textbf{\textbf{0.1504}}              &  &  & \textbf{22.83}                     & \textbf{\textbf{0.802}}            & \textbf{\textbf{0.2036}}              &  &  & \textbf{28.46}                     & \textbf{\textbf{0.891}}            & \textbf{\textbf{0.1058}}              &   \\ 
\hline
2D GS (1\%)         &  & 25.01          & 0.885                   & 0.1621                   &  &  & 23.76                              & 0.648                              & 0.4172                                &  &  & 26.99                              & 0.847                              & 0.2963                                &  &  & 19.68                              & 0.644                              & 0.4416                                &  &  & 24.83                              & 0.766                              & 0.3041                                &   \\
Textured GS (1\%)   &  & \uline{25.54}  & \uline{0.892}           & \uline{0.1523}           &  &  & \uline{24.38}                      & \uline{0.675}                      & \uline{0.3710}                        &  &  & \uline{27.02}                      & \uline{0.851}                      & \uline{0.2724}                        &  &  & \uline{20.25}                      & \uline{0.677}                      & \uline{0.3954}                        &  &  & \uline{25.23}                      & \uline{0.788}                      & \uline{0.2763}                        &   \\
Ours (1\%)          &  & \textbf{26.44} & \textbf{\textbf{0.904}} & \textbf{\textbf{0.1191}} &  &  & \textbf{24.84}                     & \textbf{\textbf{0.703}}            & \textbf{\textbf{0.3247}}              &  &  & \textbf{27.46}                     & \textbf{\textbf{0.865}}            & \textbf{\textbf{0.2417}}              &  &  & \textbf{20.60}                     & \textbf{\textbf{0.698}}            & \textbf{\textbf{0.3579}}              &  &  & \textbf{26.00}                     & \textbf{\textbf{0.818}}            & \textbf{\textbf{0.2291}}              &   \\
\bottomrule[0.8pt] 
\end{tabular}
\setlength{\tabcolsep}{6pt}
% }
\caption{
Quantitative comparison with 2DGS~\cite{huang20242d} and Textured GS~\cite{chao2025textured} under varying numbers of Gaussian primitives.
Our method consistently achieves higher fidelity across different primitive budgets. (Best scores are shown in \textbf{bold}, second-best are \underline{underlined}.)
}
\label{tab:Quantitative_comparison}
\end{table*}

\noindent\textbf{Quantitative Results.}
We evaluate our method against Textured GS~\cite{chao2025textured} and 2D GS~\cite{huang20242d} under varying primitive budgets ($100\%$, $10\%$, $1\%$).
All methods follow the same 2D GS training protocol~\cite{huang20242d}.
Table~\ref{tab:Quantitative_comparison} summarizes the results; corresponding average primitive counts are provided in the appendix.
To ensure comparable parameter budgets, we match per-Gaussian texture capacity by setting $\tau_{\text{tex}}=5$ for Textured GS and $\tau_{\text{FACT}}=4$ for our method.

As shown in Table~\ref{tab:Quantitative_comparison}, our method achieves consistently higher PSNR and SSIM, and lower LPIPS across all datasets.
The advantage becomes most pronounced under low primitive budgets ($10\%$ and $1\%$), 
where FACT-GS improves PSNR by up to \textbf{+1.0 dB} and reduces LPIPS by \textbf{22\%}.
This behavior is expected: when representational capacity is limited, 
uniform texture sampling wastes parameters in smooth regions, 
whereas our frequency-aligned parameterization concentrates capacity on high-frequency structures.

We further evaluate our method ($100\%$ primitive budget) on the NeRF Synthetic~\cite{mildenhall2021nerf} 
and MipNeRF 360-Indoor~\cite{barron2022mip} datasets. 
We compare against Textured GS~\cite{chao2025textured}, SuperGS~\cite{xu2024supergaussians}, 
GsTex~\cite{rong2025gstex}, and Neural Shell~\cite{zhang2025neural}.

As shown in Table~\ref{tab:Quantitative_comparison_additional}, our method achieves the highest overall fidelity
while maintaining real-time rendering performance (in Table~\ref{tab:inference_speed}). 
In contrast, methods such as Neural Shell~\cite{zhang2025neural} improve accuracy only by increasing computation.
This indicates that our method does not rely on capacity scaling, but instead reallocates texture resolution
to high-frequency regions—yielding structural efficiency gains directly predicted by adaptive sampling theory.

\noindent\textbf{Detail Preservation Under Limited Primitive Budgets.}
Fig.~\ref{fig:results_few} compares novel view synthesis when using only $10\%$ and $1\%$ of the default 2DGS primitive count.
2D GS~\citep{huang20242d} loses fine textures entirely due to its single-color appearance model, and Textured GS~\citep{chao2025textured} produces blurred patterns because its uniform texture grid allocates capacity evenly across the surface.
In contrast, our method preserves sharp edges, printed characters, and high-frequency materials by reallocating sampling density to regions that require detail.
This confirms that frequency-aligned parameterization maintains perceptual fidelity even when representational resources are severely constrained.

\begin{figure}[t]
    \centering
    \includegraphics[width=\linewidth]{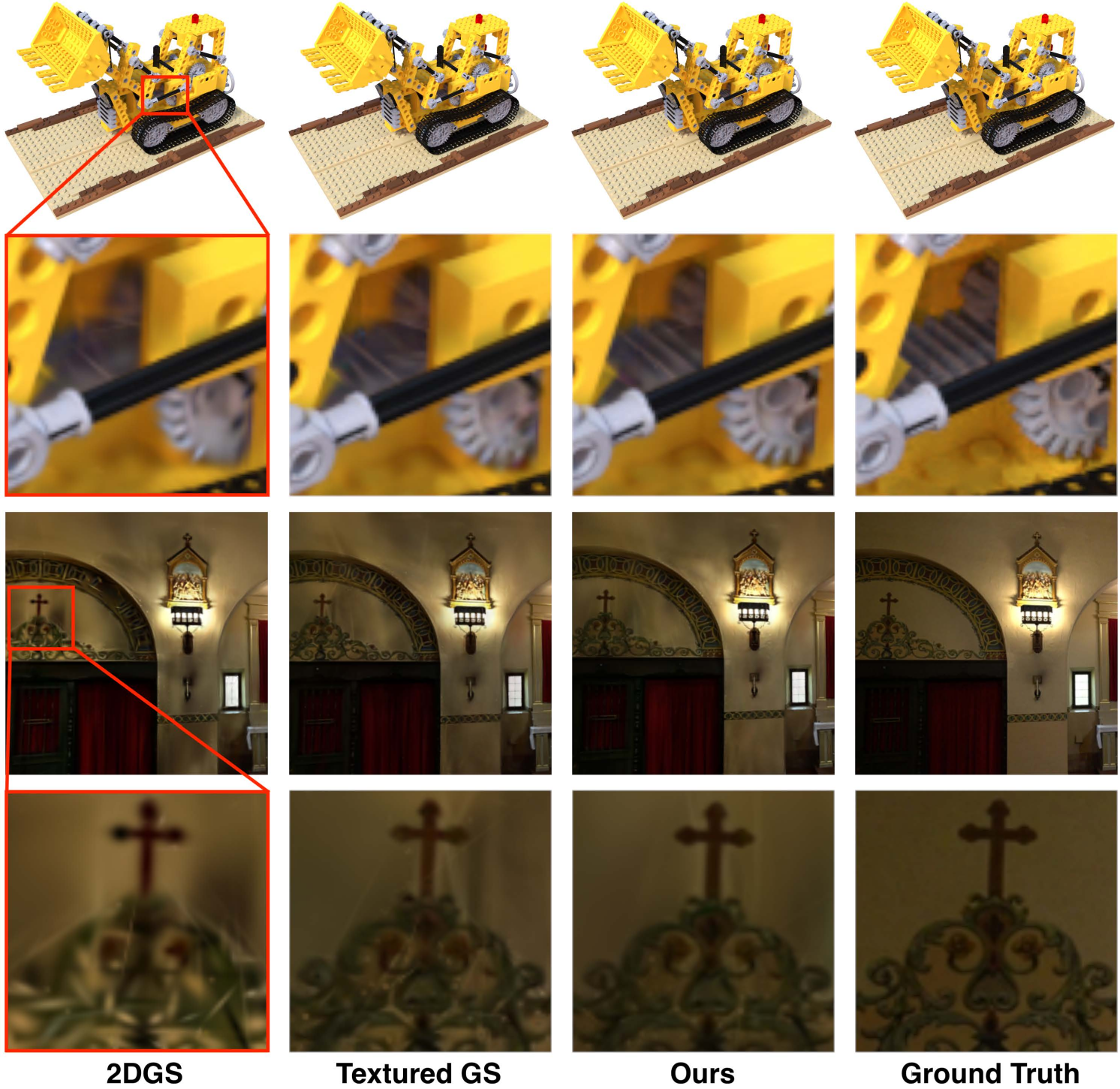}
\caption{
Qualitative comparison under high primitive budgets 
($100\%$ for \textit{Lego}, $50\%$ for \textit{Church}).
2D GS~\citep{huang20242d} lacks spatial texture variation, and Textured GS~\citep{chao2025textured}
shows blurred high-frequency patterns due to uniform texture sampling.
Our method preserves thin structures and fine texture details.
}
\label{fig:results_full}
\end{figure}

\noindent\textbf{Detail Preservation in High-Capacity Regime.}
Fig.~\ref{fig:results_full} compares reconstruction under high primitive budgets 
($100\%$ for \textit{Lego}, $50\%$ for \textit{Church}).
Even with many primitives, 2D GS loses fine texture entirely due to its single-color primitive model,
while Textured GS still blurs high-frequency patterns because its uniform texture grid
allocates texels evenly, independent of local detail.

In contrast, FACT-GS preserves thin structures, decorative patterns, and high-frequency materials.
The improvement is structural: frequency-aligned parameterization concentrates sampling density
in visually complex regions instead of wasting capacity in smooth areas.

Thus, our method improves detail fidelity even when the primitive budget is \emph{not} constrained,
demonstrating that the gain arises from better sampling allocation, not from increased capacity.

\setlength{\tabcolsep}{0.46mm} 
\begin{table}
\centering
\scriptsize
\setlength{\arrayrulewidth}{0.6pt}   % thicker rules
\renewcommand{\arraystretch}{1.12} % increase row spacing
% \resizebox{\linewidth}{!}{
\begin{tabular}{l|lcccl|lcccl} 
\toprule[0.78pt] 
\multirow{2}{*}{\textbf{Methods}}  &  & \multicolumn{3}{c}{\textbf{NeRF Synthetic}~\cite{mildenhall2021nerf}}                                             &  &  & \multicolumn{3}{c}{\textbf{MipNeRF 360-indoor}~\cite{barron2022mip}}                            &   \\ 
\cline{3-5}\cline{8-10}
                   &  & \multicolumn{1}{l}{PSNR $\uparrow$} & \multicolumn{1}{l}{SSIM $\uparrow$} & \multicolumn{1}{l}{LPIPS $\downarrow$} &  &  & PSNR $\uparrow$          & SSIM $\uparrow$                   & {LPIPS $\downarrow$}                  &   \\ 
\hline
3DGS~\cite{kerbl20233d}               &  & 33.34                    & \uline{0.969}            & 0.030                     &  &  & 31.03          & 0.921                   & 0.188                   &   \\ 

SuperGS~\cite{xu2024supergaussians}            &  & 33.71                    & \textbf{\textbf{0.970}}  & 0.031                     &  &  & 30.23          & 0.917                   & 0.188                   &   \\ 

GsTex~\cite{rong2025gstex}              &  & 33.37                    & 0.965                    & 0.041                     &  &  & 30.46          & 0.915                   & 0.204                   &   \\ 

Neural Shell~\cite{zhang2025neural}       &  & 33.50                    & 0.967                    & 0.032                     &  &  & 30.59          & 0.911                   & 0.174                   &   \\ 

Textured GS~\cite{chao2025textured} (100\%) &  & \uline{33.91}            & 0.968                    & \uline{0.024}             &  &  & \uline{31.71}  & \uline{0.938}           & \uline{0.061}           &   \\ 
\hline
Ours(100\%)        &  & \textbf{34.02}~          & \uline{0.969}            & \textbf{\textbf{0.022}}   &  &  & \textbf{31.80} & \textbf{\textbf{0.940}} & \textbf{\textbf{0.059}} &   \\
\bottomrule[0.78pt] 
\end{tabular}
\caption{Quantitative comparison with prior Gaussian-based methods, including
3DGS~\cite{kerbl20233d}, Neural Shell~\cite{zhang2025neural}, SuperGS~\cite{xu2024supergaussians},
GsTex~\cite{rong2025gstex}, and Textured GS~\cite{chao2025textured}. All methods
are evaluated using the full (100\%) 2DGS primitive budget~\cite{huang20242d} on
NeRF Synthetic~\cite{mildenhall2021nerf} and MipNeRF 360-Indoor~\cite{barron2022mip}.
(Best results are shown in \textbf{bold}, and second-best are \underline{underlined}.)}

\label{tab:Quantitative_comparison_additional}
\end{table}
\setlength{\tabcolsep}{6pt}

\begin{table}[t]
\centering
% \scriptsize
\renewcommand{\arraystretch}{1.12} % increase row spacing
\resizebox{\linewidth}{!}{
\begin{tabular}{l|c|c|c} 
\toprule[0.85pt] 
\textbf{Inference Speed (FPS) }                  & \textbf{Textured GS}~\citep{huang2024textured} & \textbf{Neural Shell}~\citep{zhang2025neural} & \textbf{Ours}  \\ 
\hline
NeRF Synthetic (100\%) & 121         & 71           & 119   \\ 
\hline
NeRF Synthetic (10\%)  & 216         & -            & 215   \\
\bottomrule[0.8pt] 
\end{tabular}
}
\caption{Inference speed (FPS$\uparrow$) comparison with Textured GS~\citep{huang2024textured} and Neural Shell~\citep{zhang2025neural} on NeRF Synthetic dataset~\cite{mildenhall2021nerf} with $100\%$ and $10\%$ default primitives of 2DGS~\cite{huang20242d}.}
\label{tab:inference_speed}
\end{table}

\subsection{Runtime and Efficiency}
As shown in Table~\ref{tab:inference_speed}, FACT-GS maintains real-time rendering
and remains within $1\text{–}2$ FPS of Textured GS~\citep{chao2025textured} on
NeRF Synthetic~\cite{mildenhall2021nerf} using a single RTX A6000 GPU. The sampling-density alignment is implemented as a per-primitive CUDA warp computed in parallel, keeping the rasterization
pipeline unchanged and thus introducing negligible overhead.

\subsection{Ablation and Analysis}

\begin{figure}[t]
    \centering
    \includegraphics[width=\linewidth]{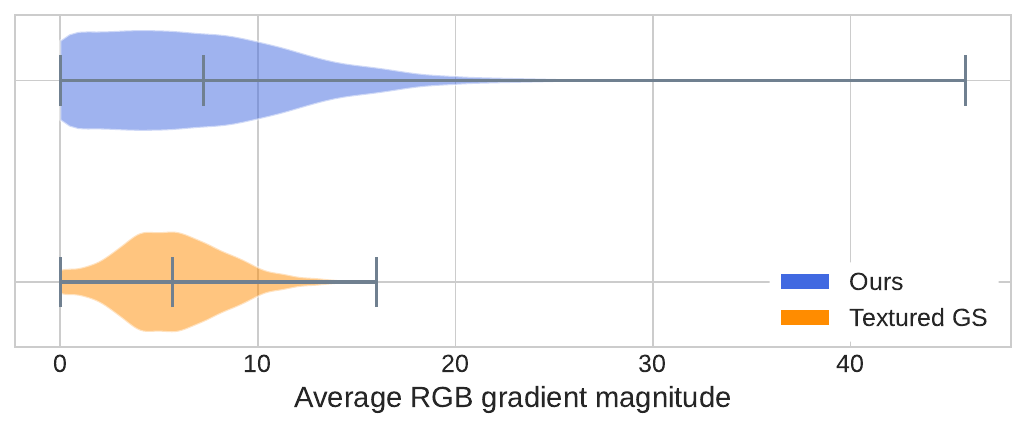}
    \caption{
    Per-Gaussian average RGB gradient magnitude. 
    Textured GS~\citep{huang2024textured} concentrates most textures in the low-frequency range ($<10$), indicating insufficient capacity for high-frequency regions. 
    Our method reallocates capacity according to local signal complexity, yielding more high-frequency textures under the same parameter budget.
    }
    \label{fig:tex_compare}
\end{figure}

\noindent\textbf{Texture Frequency Analysis.}
To quantify the effect of our reparameterization, we measure the spatial frequency of each learned
per-Gaussian RGB texture $\mathbf{T}\in\mathbb{R}^{\tau\times\tau\times 3}$ using the mean gradient magnitude~\cite{eskicioglu2002image}:
\begin{equation}
\text{Freq} = \frac{1}{3\tau^2}\sum_{c\in\{R,G,B\}}\sum_{\mathbf{p}\in\mathbf{T}} \| g(\mathbf{p}_c) \|,
\end{equation}
where $g(\cdot)$ is the Sobel gradient. 

Fig.~\ref{fig:tex_compare} shows the resulting distributions.
Visually, Textured GS~\citep{huang2024textured} forms a narrow peak in the low-frequency range ($<10$),
indicating that its uniform grid under-samples visually complex regions.
In contrast, FACT-GS produces a broader and right-shifted distribution, reflecting substantially more
high-frequency textures under the \emph{same} parameter budget.

This shows that FACT-GS does not merely sharpen appearance; it reallocates sampling capacity
toward high-frequency regions, aligning texture resolution with signal complexity.
In other words, the observed gains arise not from additional parameters, but from frequency-aligned
sampling, as predicted by our theoretical formulation.

\begin{figure}[t]
    \centering
    \includegraphics[width=\linewidth]{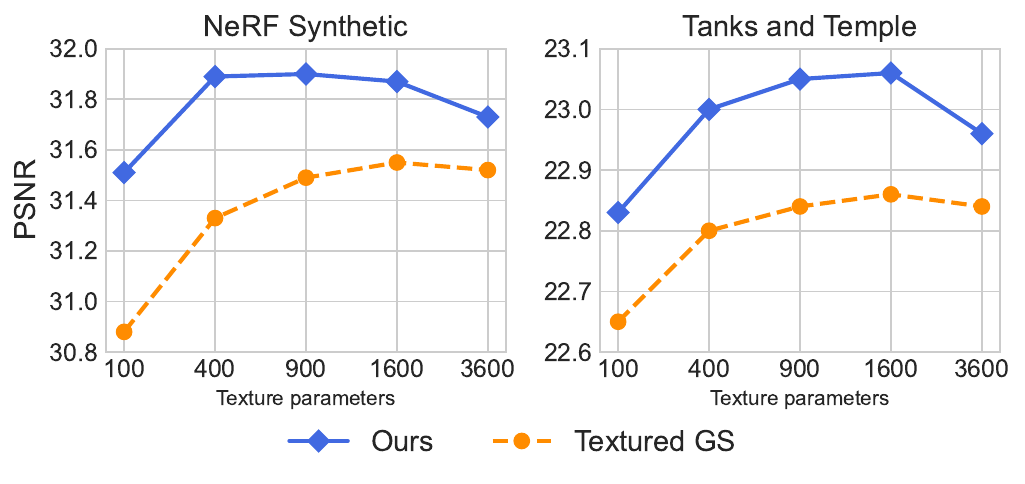}
\caption{
Ablation on the number of per-Gaussian texture parameters (PSNR, $10\%$ primitive budget).
As texture capacity increases, 
%Textured GS~\citep{huang2024textured} improves slowly and saturates at a lower PSNR,
FACT-GS maintains the lead across all parameter settings.
Our method achieves the same performance using only $\tfrac{1}{16}$ of the texture parameters.
}
    \label{fig:ablation_res}
\end{figure}

\noindent\textbf{Ablation on Texture Parameters.}
Fig.~\ref{fig:ablation_res} evaluates reconstruction quality as the number of per-Gaussian texture
parameters increases from $100$ to $3600$, using $10\%$ of the default 2DGS primitives.
Across all parameter budgets, FACT-GS achieves higher PSNR than Textured GS~\citep{huang2024textured}. 
%indicating more effective use of texture capacity.
Notably, FACT-GS matches the reconstruction quality of Textured GS while using only $\tfrac{1}{16}$ of its
texture parameters. These results confirm that frequency-aligned sampling enhances the efficiency of texture space utilization compared to uniform-grid textures, offering substantially greater improvements than simply increasing the texture resolution.

%This confirms that frequency-aligned sampling concentrates representational
%capacity in high-frequency regions, instead of allocating texels uniformly across the texture map.
%Thus, FACT-GS achieves substantially higher texture efficiency under equal or smaller parameter budgets.

\section{Conclusion}
We present FACT-GS, a frequency-aligned texture Gaussian framework that reformulates texture parameterization as an adaptive sampling problem. 
By coupling sampling density with local signal frequency through a learnable deformation field, 
%whose Jacobian modulates local area, 
FACT-GS enables adaptive, non-uniform sampling, yielding sharper details and significantly higher texture efficiency while preserving real-time rendering performance under the same parameter budget.
%under a fixed parameter budget. 

More broadly, this work shifts appearance modeling from uniform parameterization to signal-driven adaptive parameterization: assigning capacity according to information density rather than uniformly, which provides a foundation for future neural rendering methods guided by signal complexity.
%We believe this principle provides a foundation for future neural rendering representations that allocate parameters according to information rather than spatial extent.

\noindent\textbf{Limitations and Future Work.} 
FACT-GS currently only looks at the gradient of the texture and focuses on spatial alignment at the per-texture level. In the future, we are looking at incorporating explicit frequency predictors and exploring the extension of this technique to dynamic scenes.

%and extending the warping field to enforce temporal consistency in dynamic scenes. 
%and integrating neural texture fields to further enrich appearance representation.

\section*{Acknowledgment}
We acknowledge the support of the Natural Sciences and Engineering Research Council of Canada (NSERC), under funding reference numbers RGPIN-2021-03477.

{
    \small
    \bibliographystyle{ieeenat_fullname}
    \bibliography{main}
}

% WARNING: do not forget to delete the supplementary pages from your submission 
\clearpage
\setcounter{page}{1}
\maketitlesupplementary

\section{Appendix}
\subsection{Author Contributions}
Tianhao Xie and Linlian Jiang contributed equally to this work.
Tianhao Xie designed the algorithm, implemented it, and conducted the experiments.
Linlian Jiang led the method formulation and was primarily responsible for manuscript writing.
\subsection{Datasets}
We evaluate our method on five standard benchmarks for novel view synthesis, covering both synthetic and real-world scenes. 
The experimental scenes are listed below:
\begin{itemize}
    \item \textbf{NeRF Synthetic}~\citep{mildenhall2021nerf}: all scenes.
    \item \textbf{MipNeRF 360 v2}~\citep{barron2022mip}: publicly available scenes including Bicycle, Counter, Garden, Room, Bonsai, Kitchen, and Stump.
    \item \textbf{DTU}~\citep{jensen2014large}: Scan105, Scan110, Scan37, and Scan63.
    \item \textbf{Tanks \& Temples}~\citep{knapitsch2017tanks}: Train, Truck, and Church.
    \item \textbf{LLFF}~\citep{mildenhall2019llff}: Horns and Fortress.
\end{itemize}

\subsection{Budgets for Fig.~\ref{fig:results_few} and Fig.~\ref{fig:teaser}}
\label{sec:budgets}
\begin{itemize}
    \item \textbf{NeRF Ship}: $10\%$.
    \item \textbf{NeRF Chair}: $10\%$
    \item \textbf{NeRF Hotdog}: $1\%$
    \item \textbf{Tanks \& Temples Truck}: $1\%$.
    \item \textbf{Tanks \& Temples Train}: $1\%$
    \item \textbf{Tanks \& Temples Church}: $10\%$
    \item \textbf{MipNeRF 360 Kitchen}: $10\%$.
\end{itemize}

\subsection{Ablation on Texture Parameters}
\begin{figure*}[t]
    \centering
    \includegraphics[width=\linewidth]{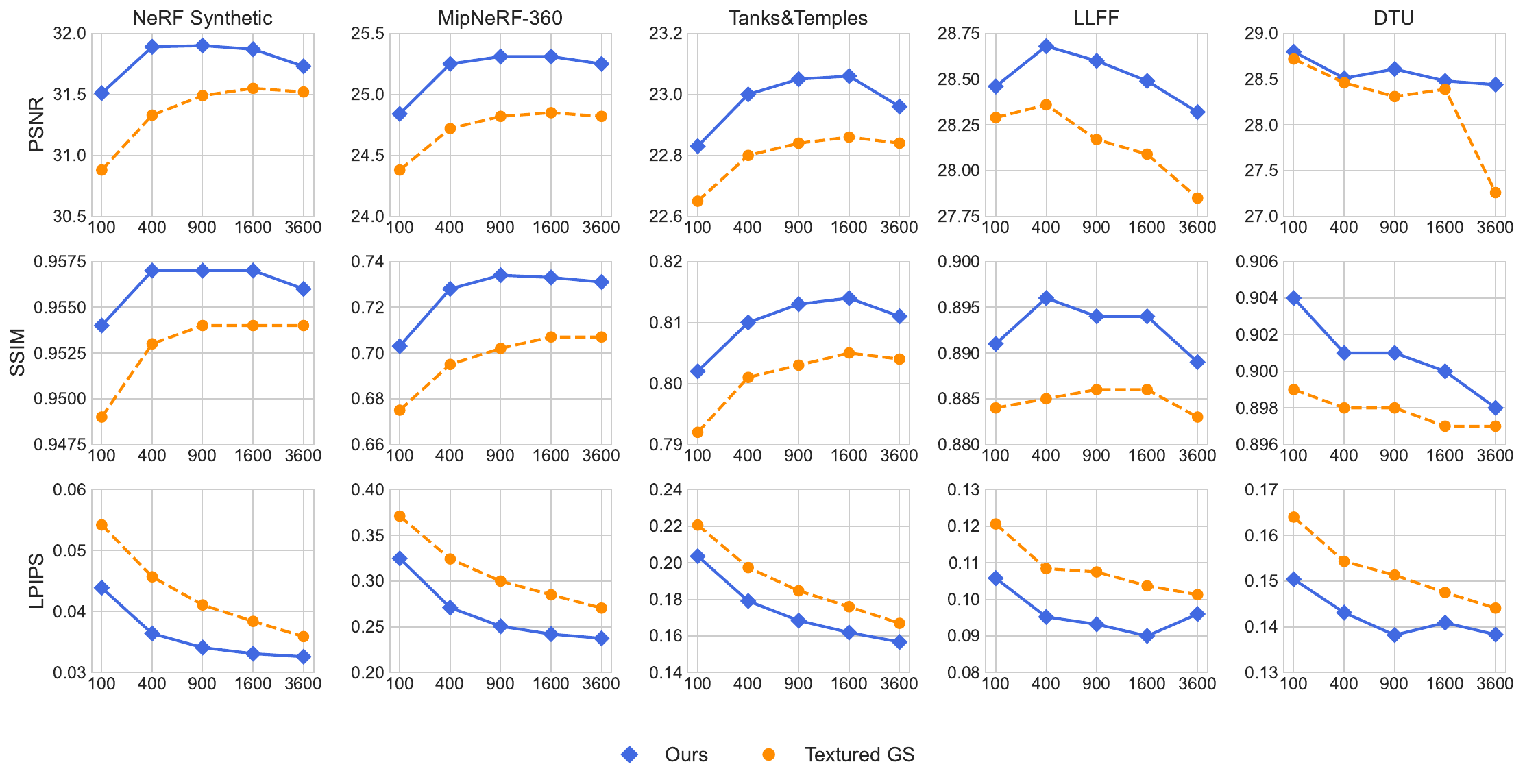}
    \vspace{-20pt}
\caption{
Ablation of the number of per-Gaussian texture parameters (the $x$ axis represents the per-Gaussian texture parameters count). The per-Gaussian texture parameters increase from $100$ to $3600$, using $10\%$ of default 2DGS primitives (the experiments on MipNeRF-360 were conducted with $1\%$ of default 2DGS primitives due to the GPU memory issue). Across all datasets and parameter settings, FACT-GS achieves a consistent lead in all metrics 
(PSNR$\uparrow$, SSIM$\uparrow$, LPIPS$\downarrow$), 
demonstrating its superior texture-space efficiency compared with uniform-grid textures.
}
\vspace{-20pt}
\label{fig:ablation_full}
\end{figure*}

We evaluated the quality of the novel view synthesis as the number of per-Gaussian texture parameters increases from $100$ to $3600$ for Textured GS and our method in 5 datasets, using $10\%$ of default 2DGS primitives (the experiments on MipNeRF-360 were conducted with $1\%$ of default 2DGS primitives due to the GPU memory issue), as shown in Fig.~\ref{fig:ablation_full}, and Table~\ref{tab:ablation_full}. The PSNR$\uparrow$, SSIM$\uparrow$, LPIPS$\downarrow$ were reported, and the $x$ axis in Fig.~\ref{fig:ablation_full} represents the per-Gaussian texture parameters count. 

Across all parameter budgets and datasets, FACT-GS achieves higher PSNR and SSIM, and lower LPIPS than Textured GS~\citep{huang2024textured}, which demonstrates that our method improves the efficiency of texture space utilization compared with uniform-grid textures, yielding substantially greater performance gains than simply increasing the texture resolution across datasets.

\noindent{\textbf{Parameters Count.}} To ensure a fair comparison with Textured GS, we matched the parameter number of both methods as closely as possible; nevertheless, our method consistently uses fewer parameters. The texture parameters of Textured GS are computed as $\tau_{\text{tex}}\times \tau_{\text{tex}}\times4$, while those of our method are $\tau_{\text{FACT}}\times \tau_{\text{FACT}}\times6$.
We report the exact number of texture parameters for both methods in Table~\ref{tab:num_parameters}.

\subsection{Model Size}
\begin{figure}[t]
    \centering
  \includegraphics[width=\linewidth]{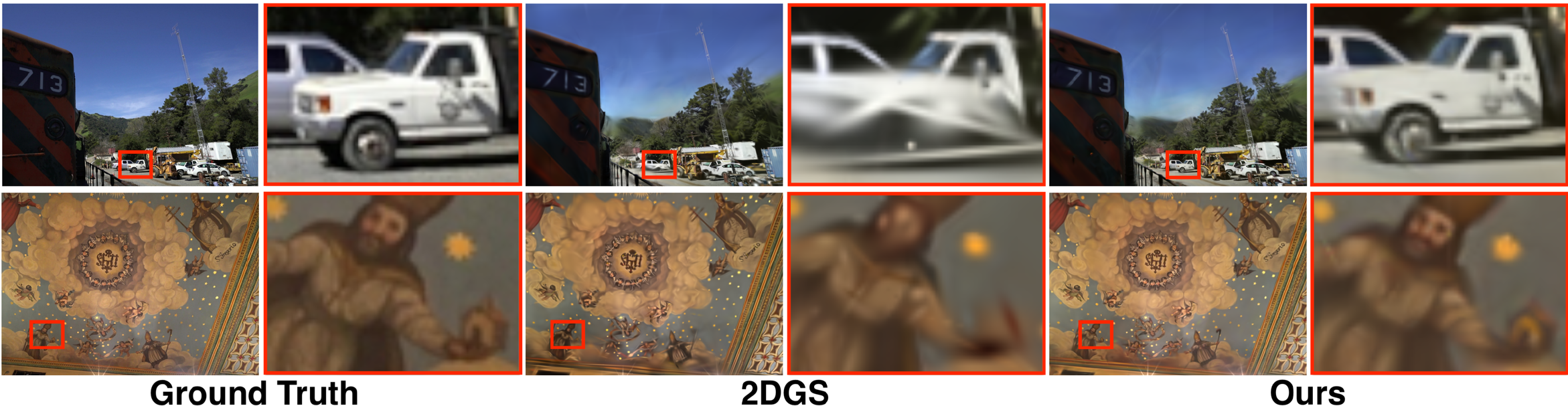}
   \vspace{-18pt}
   \caption{Quantitative size-matched comparison with 2DGS.}
   \vspace{-18pt}
    \label{fig:same_size}
\end{figure}

Model size is matched for TexturedGS and our method in the main paper with identical Gaussian counts and total texture parameter budgets; an explicit size-matched comparison with 2DGS is further reported in Tab.~\ref{tab:matched_size_comparison} and Fig.~\ref{fig:same_size}.

\subsection{Interpreting RGB Gradient.}
\begin{figure}[t]
    \centering
   \includegraphics[width=\linewidth]{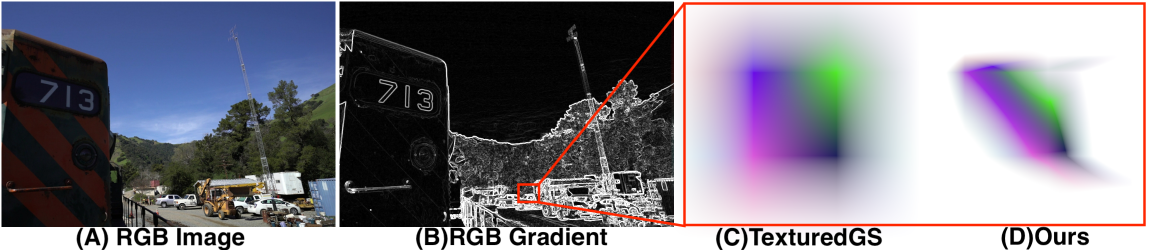}
   %\vspace{-23pt}
   \caption{RGB Gradient and per-Gaussian Texture Visualization.}
   \vspace{-8pt}
    \label{fig:freq}
\end{figure}

We provide additional qualitative evidence for texture capacity utilization in Fig.~\ref{fig:freq}. 
RGB gradient magnitude provides a practical proxy for local spatial frequency and effective texel allocation under a fixed texel budget, as higher gradients correspond to more rapidly varying color signals.
Fig.~\ref{fig:freq}(C–D) visualizes per-Gaussian local texture footprints, where our method concentrates color variation into a more compact and structured support, producing sharper transitions and clearer edges than TexturedGS. Accordingly, Fig.\ref{fig:tex_compare} shows a right-shift in the RGB gradient distribution under parameter-matched settings, indicating increased effective utilization of texture capacity for high-frequency detail reconstruction.

\begin{table}[t]
\centering
\resizebox{0.65\linewidth}{!}{
\begin{tabular}{l|l} 
\toprule
Textured GS & Ours  \\ 
\hline
$5\times5\times4 = 100$      & $4\times4\times6 = 96$    \\
$10\times10\times4 = 400$    & $8\times8\times6 = 384$    \\
$15\times15\times4 = 900$    & $12\times12\times6 = 864$ \\
$20\times20\times4 = 1600$    & $16\times16\times6 = 1536$  \\
$30\times30\times4 = 3600$    & $24\times24\times6 = 3456$  \\
\bottomrule
\end{tabular}}
\caption{Exact texture parameters count of Textured GS and our method for the ablation results in Fig.~\ref{fig:ablation_full}.}
\vspace{-20pt}
\label{tab:num_parameters}
\end{table}

\noindent\textbf{Number of Primitives.}
% We report the average $100\%$ default 2DGS primitive count of each dataset in table~\ref{tab:num_primitives}.
We report the average number of primitives in the default 100\% 2DGS setting in Table~\ref{tab:num_primitives}.

\begin{table}
\centering

\resizebox{1\linewidth}{!}{%
\begin{tabular}{l|c|lclcl} 
\toprule
\multirow{2}{*}{Method} & Average Size   &  & Tanks~\&Temples &                                              & LLFF                                                                                      & \multicolumn{1}{c}{DTU}                                                                    \\ 
\cline{3-7}
                        & (MB)   & \multicolumn{5}{c}{PSNR~($\uparrow$)~~/~~SSIM~($\uparrow$)~~/~~LPIPS~($\downarrow$)}                                                                                                                                                                      \\ 
\hline
2DGS                    & 100.94 & \multicolumn{3}{c}{22.79~/~0.805~/~0.194}                            & 27.68~/~0.881~/~0.113                                                                         & 27.85~/~0.905~/~0.133                                                                          \\
Ours                    & 99.23 & \multicolumn{3}{c}{\textbf{23.11~}/\textbf{~0.816~}/\textbf{~0.177}} & \textbf{\textbf{\textbf{\textbf{28.46~/}}\textbf{\textbf{~0.891~/}}\textbf{\textbf{~0.106}}}} & \textbf{\textbf{\textbf{\textbf{28.90~/}}\textbf{\textbf{~0.910~/}}\textbf{\textbf{~0.128}}}}  \\
\bottomrule
\end{tabular}}
\vspace{-10pt}
\caption{Size-matched comparison with 2DGS.}
\label{tab:matched_size_comparison}
\vspace{-13pt}
\end{table}

\subsection{Backpropagation of FACT Texture}

\noindent\textbf{Forward.} 
Let $\mathbf{T}_i \in \mathbb{R}^{\tau\times\tau\times4}$ denote the texture map of the $i$-th Gaussian, 
and $\mathbf{D}_i \in \mathbb{R}^{\tau\times\tau\times2}$ its learned deformation field. 
Given a camera ray from pixel $\mathbf{p}$ that intersects the $i$-th Gaussian at local coordinates $(u,v)$, 
the sampled displacement $(\Delta u, \Delta v)$ is computed by bilinear interpolation over $\mathbf{D}_i$:
\vspace{-5pt}
\begin{equation}
\label{eq:bilinear}
\Delta u = \sum_{j\in N_{(u,v)}} b_j(u,v)\, D_j^u, \qquad 
\Delta v = \sum_{j\in N_{(u,v)}} b_j(u,v)\, D_j^v,
\end{equation}
where $N_{(u,v)}$ denotes the four neighboring texels surrounding $(u,v)$, 
and $b_j(u,v)$ are their corresponding bilinear interpolation weights. 
Let $\Phi(u,v) = (u,v) + (\Delta u,\Delta v)$ be the frequency-aligned sampling coordinate. 
The sampled color $c$ from texture $\mathbf{T}_i$ is then obtained as
\vspace{-5pt}
\begin{equation}
c = \sum_{k\in N_{\Phi(u,v)}} b_k(\Phi(u,v))\, T_k,
\end{equation}
where $N_{\Phi(u,v)}$ are the four neighboring texels around $\Phi(u,v)$ and $b_k(\Phi(u,v))$ are the corresponding interpolation weights.

\noindent\textbf{Backward.} 
The gradient of the sampled color $c$ with respect to $D_j^u$ at texel $j$ is given by
\vspace{-5pt}
\begin{equation}
\label{eq:backprop}
\frac{\partial c}{\partial D_j^u}
= 
\Big(
\sum_{k\in N_{\Phi(u,v)}} T_k \, 
\frac{\partial b_k(\Phi(u,v))}{\partial (u+\Delta u)}
\Big)
\frac{\partial \Delta u}{\partial D_j^u}.
\end{equation}
Substituting Eq.~(\ref{eq:bilinear}) into Eq.~(\ref{eq:backprop}) yields
\begin{equation}
\label{eq:backprop1}
\frac{\partial c}{\partial D_j^u}
=
\Big(
\sum_{k\in N_{\Phi(u,v)}} 
T_k \,
\frac{\partial b_k(\Phi(u,v))}{\partial (u+\Delta u)}
\Big)
b_j(u,v),
\end{equation}
which defines the differentiable gradient flow from the rendered color $c$ to the deformation field $\mathbf{D}_i$.

\begin{table}[t]
\centering
\resizebox{0.70\linewidth}{!}{
\begin{tabular}{lc}
\toprule
\multicolumn{1}{l}{} & Number of primitives \\ 
\midrule
NeRF Synthetics~\cite{mildenhall2021nerf} & 91,396 \\
MipNeRF 360 v2~\cite{barron2022mip} & 2,763,229 \\
DTU~\cite{jensen2014large} & 316,989 \\
Tanks \& Temple~\cite{knapitsch2017tanks} & 1,584,660 \\
LLFF~\cite{mildenhall2019llff} & 490,511 \\
\bottomrule
\end{tabular}
}
\caption{The average number of primitives under the default 100\% optimization setting.}
\label{tab:num_primitives}
\end{table}

% \noindent\textbf{Backward}
% The gradient of the deformation field $D^u$ at $j^{th}$ texel is
% \begin{equation}
% \label{eq:backprop}
%     \frac{\partial c}{\partial \mathbf{D}_j^u} = (\sum_{k\in N_{\Phi(u,v)}} \mathbf{T}_k(\Phi(u,v))\cdot \frac{\partial b_k(\Phi(u,v))}{\partial (u+\Delta u)} )\cdot \frac{\partial \Delta u}{\partial D_j^u(u,v)},
% \end{equation}
% Substituting eq~\ref{eq:bilinear} into equation~\ref{eq:backprop}, we get
% \begin{equation}
% \label{eq:backprop1}
%     \frac{\partial c}{\partial \mathbf{D}_j^u} = (\sum_{k\in N_{\Phi(u,v)}} \mathbf{T}_k(\Phi(u,v))\cdot \frac{\partial b_k(\Phi(u,v))}{\partial (u+\Delta u)} )\cdot b_j(u,v).
% \end{equation}
%Equation~\ref{eq:backprop} can be applied to the deformation field $D^v$ by replacing $\frac{\partial b_j}{\partial (u+\Delta u)}$ to $\frac{\partial b_j}{\partial (v+\Delta v)}$.

% \begin{table}[t]
% \centering
% \begin{tabular}{cc} 
% \toprule
% \multicolumn{1}{l}{} & Number of primitives~  \\ 
% \hline
% NeRF Synthetics~\cite{mildenhall2021nerf}      & 91,396                 \\ 
% \hline
% MipNeRF 360 v2~\cite{barron2022mip}       & 2,763,229              \\ 
% \hline
% DTU~\cite{jensen2014large}                  & 316,989                \\ 
% \hline
% Tanks \& Temple~\cite{knapitsch2017tanks}        & 1,584,660              \\ 
% \hline
% LLFF~\cite{mildenhall2019llff}                & 490,511                \\
% \bottomrule
% \end{tabular}
% \caption{The average number of primitives under default optimization strategy ($100\%$).}
% \label{tab:num_primitives}
% \end{table}

\setlength{\tabcolsep}{0.765mm} 
\begin{table*}[t]
\centering
\scriptsize
\setlength{\arrayrulewidth}{0.6pt}   % thicker rules
\renewcommand{\arraystretch}{1.22} % increase row spacing
% \resizebox{\linewidth}{!}{
\begin{tabular}{c|lcccl|lcccl|lcccl|lcccl|lcccl} 
\toprule[0.8pt]
\multirow{2}{*}{\textbf{Methods}} & \multicolumn{1}{c}{~} & \multicolumn{3}{c}{\textbf{NeRF Synthetic}~\cite{mildenhall2021nerf}} &  & \multicolumn{1}{c}{} & \multicolumn{3}{c}{\textbf{MipNeRF 360v2}~\cite{barron2022mip}}                                                                     & \multicolumn{1}{c|}{} &  & \multicolumn{3}{c}{\textbf{DTU}~\cite{jensen2014large}}                                                                               & \multicolumn{1}{c|}{} & \multicolumn{1}{c}{} & \multicolumn{3}{c}{\textbf{Tanks \& Temples}~\cite{knapitsch2017tanks}}                                                                     &  &  & \multicolumn{3}{c}{\textbf{LLFF}~\cite{mildenhall2019llff}}                                                                              & \multicolumn{1}{c}{}  \\ 
\cline{3-5}\cline{8-10}\cline{13-15}\cline{18-20}\cline{23-25}
                                  &                       & PSNR$\uparrow$ & SSIM$\uparrow$ & LPIPS$\downarrow$  &  &                      & \multicolumn{1}{l}{PSNR$\uparrow$} & \multicolumn{1}{l}{SSIM$\uparrow$} & \multicolumn{1}{l}{LPIPS$\downarrow$} &                       &  & \multicolumn{1}{l}{PSNR$\uparrow$} & \multicolumn{1}{l}{SSIM$\uparrow$} & \multicolumn{1}{l}{LPIPS$\downarrow$} &                       &                      & \multicolumn{1}{l}{PSNR$\uparrow$} & \multicolumn{1}{l}{SSIM$\uparrow$} & \multicolumn{1}{l}{LPIPS$\downarrow$} &  &  & \multicolumn{1}{l}{PSNR$\uparrow$} & \multicolumn{1}{l}{SSIM$\uparrow$} & \multicolumn{1}{l}{LPIPS$\downarrow$} &                       \\ 
\hline
Textured GS (100)                 &                       & 30.88          & 0.949          & 0.0542             &  &                      & 24.38                              & 0.675                              & 0.3710                                &                       &  & 28.72                              & 0.899                              & 0.1640                                &                       &                      & 22.65                              & 0.792                              & 0.2206                                &  &  & 28.29                              & 0.884                              & 0.1206                                &                       \\
Ours (96)                         &                       & \textbf{31.51} & \textbf{0.954} & \textbf{0.0439}    &  &                      & \textbf{24.84}                     & \textbf{0.703}                     & \textbf{0.3247}                       &                       &  & \textbf{28.80}                     & \textbf{0.904}                     & \textbf{0.1504}                       &                       &                      & \textbf{22.83}                     & \textbf{0.802}                     & \textbf{0.2036}                       &  &  & \textbf{28.46}                     & \textbf{0.891}                     & \textbf{0.1058}                       &                       \\ 
\hline
Textured GS (400)                 &                       & 31.33          & 0.953          & 0.0457             &  &                      & 24.72                              & 0.695                              & 0.3242                                &                       &  & 28.46                              & 0.898                              & 0.1543                                &                       &                      & 22.80                              & 0.801                              & 0.1974                                &  &  & 28.36                              & 0.885                              & 0.1084                                &                       \\
Ours (384)                        &                       & \textbf{31.89} & \textbf{0.957} & \textbf{0.0364}    &  &                      & \textbf{25.25}                     & \textbf{0.728}                     & \textbf{0.2710}                       &                       &  & \textbf{28.51}                     & \textbf{0.901}                     & \textbf{0.1431}                       &                       &                      & \textbf{23.00}                     & \textbf{0.810}                     & \textbf{0.1791}                       &  &  & \textbf{28.68}                     & \textbf{0.896}                     & \textbf{0.0952}                       &                       \\ 
\hline
Textured GS (900)                 &                       & 31.49          & 0.954          & 0.0411             &  &                      & 24.82                              & 0.702                              & 0.3000                                &                       &  & 28.31                              & 0.898                              & 0.1513                                &                       &                      & 22.84                              & 0.803                              & 0.1847                                &  &  & 28.17                              & 0.886                              & 0.1075                                &                       \\
Ours (864)                        &                       & \textbf{31.90} & \textbf{0.957} & \textbf{0.0341}    &  &                      & \textbf{25.31}                     & \textbf{0.734}                     & \textbf{0.2506}                       &                       &  & \textbf{28.61}                     & \textbf{0.901}                     & \textbf{0.1382}                       &                       &                      & \textbf{23.05}                     & \textbf{0.813}                     & \textbf{0.1684}                       &  &  & \textbf{28.60}                     & \textbf{0.894}                     & \textbf{0.0932}                       &                       \\ 
\hline
Textured GS (1600)                &                       & 31.55          & 0.954          & 0.0383             &  &                      & 24.85                              & 0.707                              & 0.2850                                &                       &  & 28.39                              & 0.897                              & 0.1475                                &                       &                      & 22.86                              & 0.805                              & 0.1760                                &  &  & 28.09                              & 0.886                              & 0.1037                                &                       \\
Ours (1536)                       &                       & \textbf{31.87} & \textbf{0.957} & \textbf{0.0331}    &  &                      & \textbf{25.31}                     & \textbf{0.733}                     & \textbf{0.2420}                       &                       &  & \textbf{28.48}                     & \textbf{0.900}                     & \textbf{0.1409}                       &                       &                      & \textbf{23.06}                     & \textbf{0.814}                     & \textbf{0.1619}                       &  &  & \textbf{28.49}                     & \textbf{0.894}                     & \textbf{0.0900}                       &                       \\ 
\hline
Textured GS (3600)                &                       & 31.52          & 0.954          & 0.0359             &  &                      & 24.82                              & 0.707                              & 0.2705                                &                       &  & 27.26                              & 0.897                              & 0.1441                                &                       &                      & 22.84                              & 0.804                              & 0.1669                                &  &  & 27.85                              & 0.883                              & 0.1013                                &                       \\
Ours (3456)                       &                       & \textbf{31.73} & \textbf{0.956} & \textbf{0.0326}    &  &                      & \textbf{25.25}                     & \textbf{0.731}                     & \textbf{0.2374}                       &                       &  & \textbf{28.44}                     & \textbf{0.898}                     & \textbf{0.1383}                       &                       &                      & \textbf{22.96}                     & \textbf{0.811}                     & \textbf{0.1567}                       &  &  & \textbf{28.32}                     & \textbf{0.889}                     & \textbf{0.0960}                       &                       \\
\bottomrule[0.8pt]
\end{tabular}
\setlength{\tabcolsep}{6pt}
% }
\caption{
Ablation on the number of per-Gaussian texture parameters.
In the Methods column, the numbers in parentheses indicate the exact texture parameter counts.
We report PSNR$\uparrow$, SSIM$\uparrow$, and LPIPS$\downarrow$, which are the metrics used to plot the curves in Fig.~\ref{fig:ablation_full}.
Best results are highlighted in \textbf{bold}.
}
\label{tab:ablation_full}
\end{table*}

%Given an image coordinate $\mathbf{x}=(x,y)$, 
%the intersection between the viewing ray and the Gaussian’s tangent plane 
%is computed as the intersection of two orthogonal image-space planes.
%We represent them in homogeneous form as $\mathbf{h}_x=(-1,0,0,x)^{\top}$ and $\mathbf{h}_y=(0,-1,0,y)^{\top}$, 
%Transforming these planes into the Gaussian’s local frame gives
%\begin{equation}
%    \mathbf{h}_{\beta} = (WH)^{\!\top}\mathbf{h}_x, \quad
%    \mathbf{h}_{\gamma} = (WH)^{\!\top}\mathbf{h}_y,
%\end{equation}
%where $W$ is the world-to-screen projection and $H$ the local tangent transform.
%The local intersection coordinates $(\beta,\gamma)$ are then obtained by solving 
%a two-plane system:
%\begin{equation}
%\small
%[\beta,\gamma]^{\!\top} 
%= \left( 1\,/\!( h_\beta^1h_\gamma^2 - h_\beta^2h_\gamma^1) \right)
%\begin{bmatrix}
%    h_\beta^2h_\gamma^4 - h_\beta^4h_\gamma^2 \\[3pt]
%    h_\beta^4h_\gamma^1 - h_\beta^1h_\gamma^4
%\end{bmatrix}.
%\end{equation}
%These local coordinates $\mathbf{s}(\mathbf{x})=(\beta,\gamma)$ 
%serve as the texture sampling points for each Gaussian.

\end{document}